\documentclass[conference]{IEEEtran}
\IEEEoverridecommandlockouts
\usepackage{cite}
\usepackage{amsmath,amssymb,amsfonts}
\usepackage{algorithmic}
\usepackage{graphicx}
\usepackage{textcomp}
\usepackage{booktabs}
\usepackage{xcolor}
\def\BibTeX{{\rm B\kern-.05em{\sc i\kern-.025em b}\kern-.08em
    T\kern-.1667em\lower.7ex\hbox{E}\kern-.125emX}}
\usepackage{enumitem}
    
\usepackage{siunitx}
\usepackage{tikz}
\usepackage{blindtext}
\usepackage{hyperref}
\usepackage{multirow}
\usepackage{nicefrac}
\usepackage{placeins}

\usepackage{lipsum}

\usepackage{dblfloatfix}

\usepackage{fancyhdr}

\newcommand{\uproman}[1]{\uppercase\expandafter{\romannumeral#1}}    
    
\newcommand{\discolorlinks}[1]{{\hypersetup{hidelinks}#1}}
    
\begin{document}

\title{Efficient Multi-Task Scene Analysis with RGB-D Transformers

\thanks{\footnotesize This work has received funding from Carl-Zeiss-Stiftung to the project \textit{Co-Presence of Humans and Interactive Companions for Seniors} (CO-HUMANICS).}
}

\author{\IEEEauthorblockN{Söhnke Benedikt Fischedick, Daniel Seichter, Robin Schmidt, Leonard Rabes, and Horst-Michael Gross}
\IEEEauthorblockA{\textit{Ilmenau University of Technology, Neuroinformatics and Cognitive Robotics Lab} \\
98684 Ilmenau, Germany \\
\small\tt soehnke.fischedick@tu-ilmenau.de, ORCID:
\discolorlinks{\href{https://orcid.org/0000-0001-8447-0584}{0000-0001-8447-0584}}}
}

\maketitle

\newboolean{isarxiv}
\setboolean{isarxiv}{true}
\ifthenelse{\boolean{isarxiv}}{%
    \renewcommand{\headrulewidth}{0pt}
    \fancypagestyle{fancyfirstpage}{%
                \fancyhf{}%
        \fancyhead[C]{%
            \footnotesize%
            \textcolor{gray}{%
                © 2023 IEEE.
                Personal use of this material is permitted.
                Permission from IEEE must be obtained for all other uses, in any current or future media, including reprinting/republishing this material for advertising or promotional purposes, creating new collective works, for resale or redistribution to servers or lists, or reuse of any copyrighted component of this work in other works.
                DOI:\href{https://doi.org/10.1109/IJCNN54540.2023.10191977}{10.1109/IJCNN54540.2023.10191977}
            }%
        }%
        \fancyfoot[C]{%
            \footnotesize%
            \textcolor{gray}{\thepage}%
        }
    }
    \fancypagestyle{fancypage}{%
        \fancyhf{}%
        \fancyfoot[C]{%
            \footnotesize%
            \textcolor{gray}{\thepage}%
        }        
    }    
    \thispagestyle{fancyfirstpage}
    \pagestyle{fancypage}
}{%
    \thispagestyle{empty}%
    \pagestyle{empty}%
}%

\begin{abstract}
Scene analysis is essential for enabling autonomous systems, such as mobile robots, to operate in real-world environments.
However, obtaining a comprehensive understanding of the scene requires solving multiple tasks, such as panoptic segmentation, instance orientation estimation, and scene classification.
Solving these tasks given limited computing and battery capabilities on mobile platforms is challenging.
To address this challenge, we introduce an efficient multi-task scene analysis approach, called EMSAFormer, that uses an RGB-D Transformer-based encoder to simultaneously perform the aforementioned tasks.
Our approach builds upon the previously published EMSANet.
However, we show that the dual CNN-based encoder of EMSANet can be replaced with a single Transformer-based encoder.
To achieve this, we investigate how information from both RGB and depth data can be effectively incorporated in a single encoder.
To accelerate inference on robotic hardware, we provide a custom NVIDIA TensorRT extension enabling highly optimization for our EMSAFormer approach.
Through extensive experiments on the commonly used indoor datasets NYUv2, SUNRGB-D, and ScanNet, we show that our approach achieves state-of-the-art performance while still enabling inference with up to 39.1$\,$FPS on an NVIDIA Jetson AGX Orin 32$\,$GB.
\end{abstract}

\section{Introduction}
\label{sec:introduction}
\begin{figure}[!t]
    \vspace{0.5mm}
	\centering
	\includegraphics[width=\columnwidth]{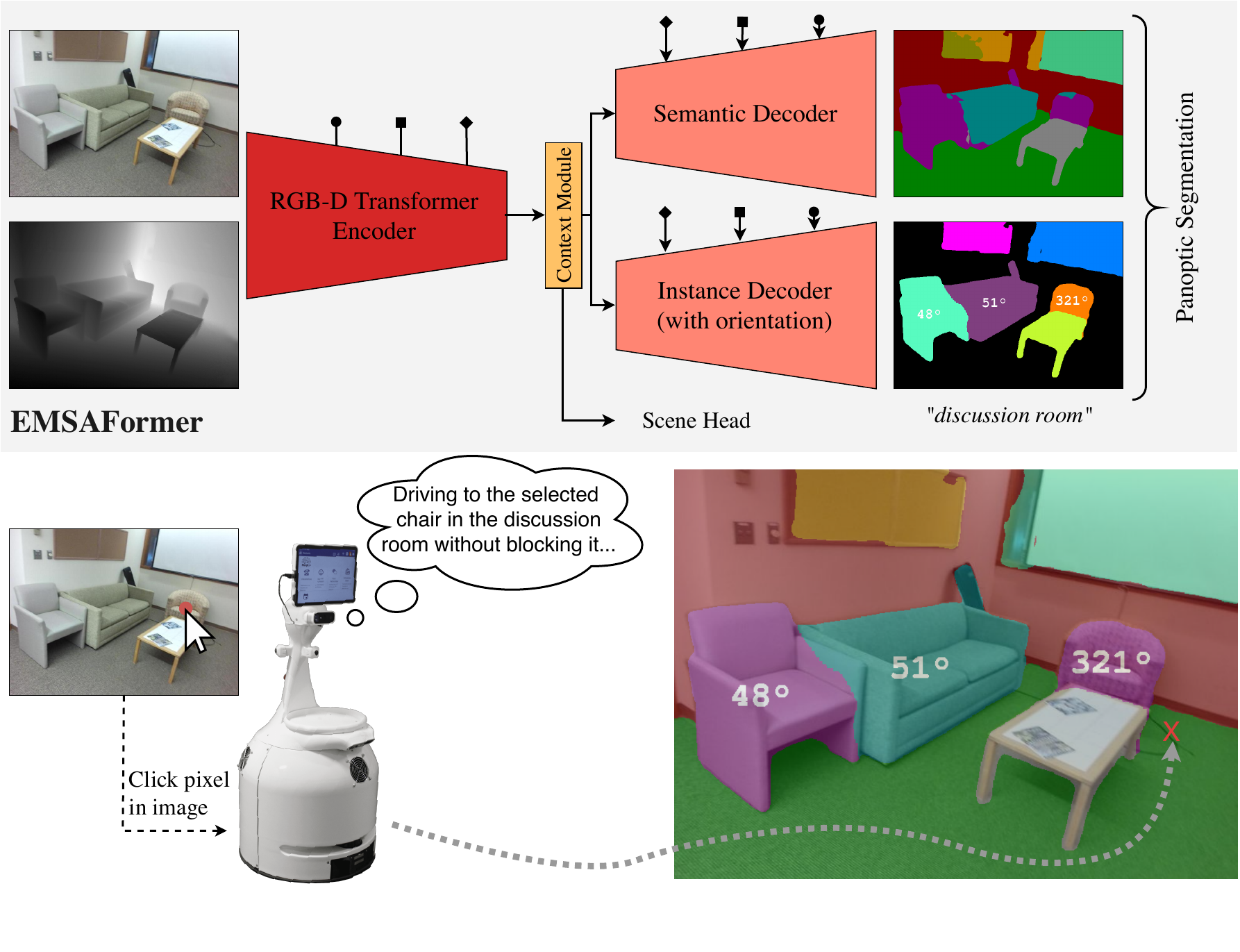}
	\vspace{-12.5mm}
	\caption{%
	    Application~(bottom) of our proposed efficient multi-task scene analysis approach with an RGB-D Transformer encoder, called EMSAFormer, that simultaneously performs panoptic segmentation, instance orientation estimation, and scene classification~(top).
        See Fig.~\ref{fig:architecture} for prediction colors.
	}
	\label{fig:eyecatcher}
	\vspace{-5mm}
\end{figure}

A broad understanding of the scene is crucial for mobile agents to operate autonomously in indoor environments.
Traditional approaches often only provide semantic or instance information, which is not sufficient for many real-world applications.
For example, in our research projects \href{http://co-humanics.de/}{CO-HUMANICS} and \href{http://morphia-projekt.de/}{MORPHIA}~\cite{morphia-isr2022}, a mobile robot should autonomously operate in indoor environments and should also be controlled by an operator from a remote location.
To enable such a remote control to inexperienced operators, a more intuitive way for navigating is required.
As shown in Fig.~\ref{fig:eyecatcher}, the operator should be able to click onto a specific point or object visible in the camera image of the current surrounding, to which the robot should automatically navigate to.
For example, the mobile robot should drive to a chair within a group of many chairs while respecting the chair's orientation to not block it. %
Furthermore, the operator should be able to select an entire room to which the robot should drive to.
To achieve this, a broader scene understanding is required.
The robot needs to combine information from multiple tasks, i.e., semantic and instance segmentation (panoptic segmentation), instance orientation estimation, and scene classification.
Due to the limited computing and battery capabilities of a mobile robot, a multi-task approach should be preferred.

To gather information for all aforementioned tasks, we introduce an efficient multi-task scene analysis approach utilizing a Transformer-based encoder~(EMSAFormer).
It builds on top of our Efficient Multi-task Scene Analysis Network~(EMSANet)~\cite{emsanet2022ijcnn} that already realizes such a system.
In~\cite{emsanet2022ijcnn}, we have shown the effectiveness of using RGB and depth as complementary data to improve the performance of individual tasks.
For processing the different modalities, a dual ResNet34 encoder is used.
The approach enables real-time application~(in our application scenario $\ge$20$\,$FPS) on an NVIDIA Jetson AGX Xavier, while still reaching state-of-the-art performance.
With the release of the NVIDIA Jetson AGX Orin 32$\,$GB, computing capabilities have been increased, i.e., it is now possible to apply larger models in real time.
Compared to NVIDIA Jetson AGX Xavier, inference throughput has almost doubled with the same power consumption.
However, the authors of EMSANet have shown that a larger encoder, e.g., utilizing two ResNet101 backbones, only slightly improves performance.
Motivated by the upcoming usage of Transformer-based architectures for computer vision, we address this limitation by replacing the dual encoder based on convolutional neural networks~(CNNs) with a single Swin Transformer for processing RGB-D data.
We conduct detailed experiments to show the effectiveness of using a Swin Transformer for processing RGB and depth data and how it compares to the traditional CNN-based encoders.
Furthermore, we address differences and challenges when incorporating a Transformer-based encoder.
Our experiments show that our EMSAFormer approach is able to outperform the state-of-the-art method EMSANet in most tasks.
Finally, to enable inference in real time on the NVIDIA Jetson AGX Orin on our mobile robot, we provide a custom NVIDIA TensorRT extension that greatly accelerates inference.
In summary, our main contributions include:
\begin{itemize}[leftmargin=5mm]
\item A complementary RGB-D encoder approach that replaces the dual encoder in EMSANet with a novel single Swin Transformer encoder, while still effectively incorporating information from both RGB and depth data due to a modified Transformer design.
\item A custom NVIDIA TensorRT extension for inference acceleration that enables using Swin Transformers as a general-purpose backbone for downstream tasks with import from Open Neural Network Exchange (ONNX) format and arbitrary input resolution.
\item Detailed quantitative and qualitative experiments on the common indoor datasets NYUv2~\cite{NYUv2-eccv2012}, SUNRGB-D~\cite{SUNRGBD-cvpr2015}, and ScanNet~\cite{scannet-cvpr2017} demonstrating the applicability and state-of-the-art performance of our approach.
\end{itemize}

Our code as well as the network weights are publicly available at: {\small\href{https://github.com/TUI-NICR/EMSAFormer}{\texttt{\url{https://github.com/TUI-NICR/EMSAFormer}}}}.

\section{Related Work}
\label{sec:related_work}
In the following, we summarize related work for scene analysis with focus on Transformer-based encoders and RGB-D processing.
We give a brief introduction how the individual tasks can be accomplished in an encoder-decoder approach.

\subsection{Transformer-based Encoders}
In recent years, convolutional neural networks~(CNNs) have been the dominant approach for various computer vision tasks, such as image classification, object detection, or semantic segmentation.
Various backbones, such as ResNet~\cite{ResNet-cvpr2016}, EfficientNet~\cite{EfficientNet-icml2019}, or ConvNeXt~\cite{convnext-cvpr2022} have been proposed and achieve state-of-the-art performance on a variety of benchmarks.
Inspired by the success of Transformers for natural language processing~(NLP)~\cite{Transformer-neurips2017, bert-arxiv2018, gpt-neurips2017}, the work of \cite{vit-iclr2021} introduces the Vision Transformer~(ViT), which is able to reach similar performance to CNNs for image classification.
While the proposed architecture achieves impressive results, it still requires extensive pretraining for comparable performance on the ImageNet benchmark~\cite{ImageNet-ijcv2015} and does not realize a pyramid structure, making it less suitable as a general-purpose backbone~\cite{swin-iccv2021, SegFromer-neurips2021}.
Motivated by the success of ViT, various Transformer-based architectures, such as Pyramid Vision Transformer~\cite{pvt-iccv2021}, Swin Transformers~\cite{swin-iccv2021, swinv2-cvpr2022}, and SegFormer~\cite{SegFromer-neurips2021} have been introduced, improving data efficiency and enabling the usage as general-purpose backbone due to the pyramid-like structure.
These architectures achieve state-of-the-art performance on a variety of benchmarks; however, they have been proposed for processing RGB data and have not yet been extended to processing RGB-D data, which is the focus of this work.

\subsection{RGB-D Encoders}
Combining RGB and depth data can improve performance in various computer vision tasks as both provide complementary features~\cite{FuseNet-accv2016, RedNet-arxiv2018, ACNet-icip2019}. 
RGB data provide information about semantic features, such as object color and texture, while depth data provide geometric information about the scene.
Handling multiple modalities can be challenging as they contain different features with deviating statistics and characteristics.
Thus, multiple approaches have been emerged.

The majority of approaches~\cite{FuseNet-accv2016, RDFNet-iccv2017, RedNet-arxiv2018, CouplingTwoStream-icip2019, SSMA-ijcv2019, MMAF-Net-arxiv2019, SA-Gate-eccv2020, esanet2021icra, emsanet2022ijcnn} handles both modalities in two separated encoders and fuse features using a dedicated fusion mechanism.
The fusion is mostly done after each resolution stage~\cite{FuseNet-accv2016, LDFNet-icip2019, RedNet-arxiv2018, esanet2021icra, emsanet2022ijcnn} of the encoders, at the end of the encoders~\cite{lstmcf-eccv2016}, or into a third encoder handling joint features~\cite{MFNet-iros2017, MultiLevelRGBD-icip2017, ACNet-icip2019, SSMA-ijcv2019}.
Moreover, most approaches additionally fuse features from the encoder to the decoder~\cite{RedNet-arxiv2018, esanet2021icra, emsanet2022ijcnn} similar to DeepLabV3+~\cite{DeepLabv3plus-eccv-2018}.
While enabling independent processing of both modalities, this method implies a crucial design decision, i.e., it introduces the difficulty of deciding where the fusion should happen and how the features are fused.
Additionally, depending on the used backbones, a dual-encoder design often leads to increased computational cost, making these approaches less suitable for embedded hardware.

Other approaches try to handle both modalities in a single encoder~\cite{DepthAwareCNN-eccv2018, 2.5DConvolutionRGBD-icip2019, Malleable2.5D-eccv2020, SGNet-arxiv2020, shapeconv-iccv2021}.
This is done by using specially tailored convolutions for incorporating depth data.
However, such approaches often lack optimization and, thus, are less suitable for embedded hardware.

Motivated by the performance achieved by Swin Transformers~\cite{swin-iccv2021,swinv2-cvpr2022}, the authors of OMNIVORE~\cite{omnivore-cvpr2022} propose a method for handling multiple modalities in a single encoder.
However, as the approach mainly focuses on handling many different modalities and large backbones, it again lacks optimization and efficiency.
Moreover, OMNIVORE requires heavy pretraining to achieve good performance.

In this paper, we follow the recent trend of using Transformer-based architectures.
However, in contrast to the aforementioned approaches, we focus on modifying the Swin Transformer architecture to efficiently incorporate depth information while still relying on a single encoder.

\begin{figure*}[!b]%
    \centering%
    \vspace{-6mm}%
        \resizebox{0.999\textwidth}{!}{%
            \begin{tikzpicture}[every node/.style={inner sep=0,outer sep=0}]%
                \node[anchor = north east] at(0, 0){%
                    \includegraphics[scale=0.5, trim=0mm 1mm 0mm 0mm, clip]{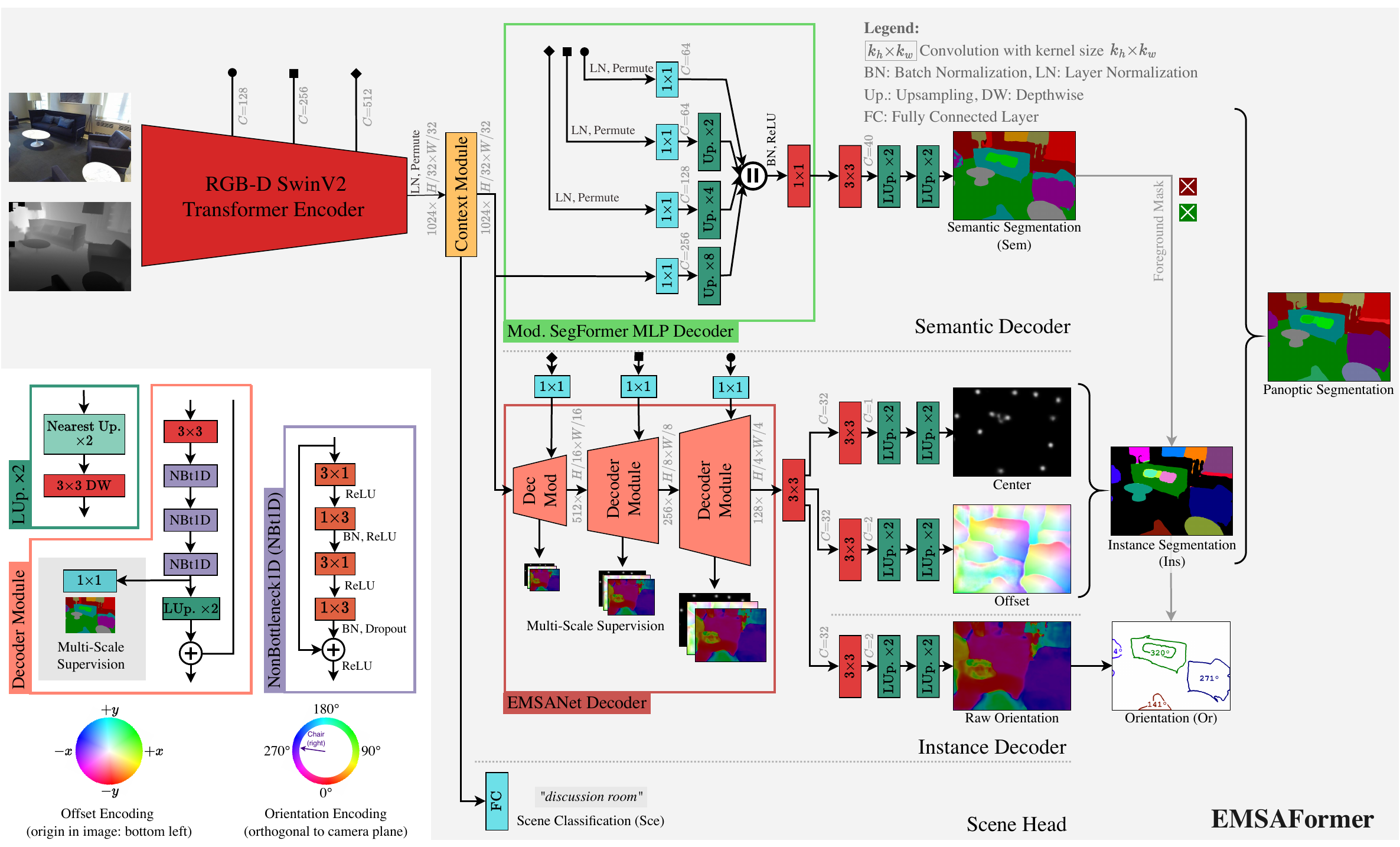}%
                };%
            \end{tikzpicture}%
        }%
    \vspace{-3mm}%
    \caption{%
        Architecture of our proposed efficient multi-task scene analysis approach with a single RGB-D Transformer encoder~(EMSAFormer) that simultaneously performs panoptic segmentation, instance orientation estimation, and scene classification. %
        For further details and explanations, see Sec.~\ref{sec:main}.
        Semantic colors are chosen as in~\cite{emsanet2022ijcnn} and are \href{https://github.com/TUI-NICR/nicr-scene-analysis-datasets/blob/v0.5.3/nicr_scene_analysis_datasets/datasets/nyuv2/nyuv2.py\#L193}{the default colors for NYUv2}~\cite{NYUv2-eccv2012}.
        Panoptic is visualized by small color differences.
    }%
    \label{fig:architecture}%
\end{figure*}

\subsection{Task Decoders}
The design of the decoder depends on the task to solve.
Scene classification, i.e., assigning a scene label, such as living room or office, to the entire input, is similar to other classification tasks.
It only requires a classification layer at the end of the encoder.
By contrast, pixel-wise dense prediction tasks require more sophisticated decoders.
As the encoder typically lowers the spatial resolution, dense-prediction decoders often incorporate multiple modules to gradually restore the resolution.
Most approaches use a CNN-based decoder~\cite{PSPNet-cvpr2017, DeepLabv3-arxiv2017, DeepLabv3plus-eccv-2018, SwiftNet-cvpr2019, esanet2021icra} of varying complexity.
With the rise of Transformer-based architectures, SegFormer~\cite{SegFromer-neurips2021} proposes another more lightweight MLP-based decoder.
Features from different stages are embedded using a fully-connected layer, upsampled to the same spatial resolution, concatenated, and passed through two additional fully-connected layers to encode the final prediction.

For panoptic segmentation, at least one dense-prediction decoder is required.
Panoptic segmentation~\cite{Panopticsegmentation-cvpr2019} combines semantic and instance segmentation and aims at assigning a semantic class to each pixel of the input image as well as a unique instance ID to each pixel belonging to a distinguishable instance.
Panoptic segmentation can be done in a top-down, bottom-up, or end-to-end manner.
Top-Down methods are typically based on existing approaches for instance segmentation and extend them with an additional decoder for semantic segmentation~\cite{PanopticFPN-cvpr2019, UPSNet-cvpr2019}.
While achieving state-of-the-art performance, these architectures typically feature sophisticated network architectures and require further logic to resolve overlapping instance masks.
Bottom-up methods~\cite{DeeperLab-arxiv2019, PanopticDeeplab-cvpr2020, emsanet2022ijcnn}, on the other hand, are often based on encoder-decoder architectures for semantic segmentation and extend them by an additional decoder for instances.
This additional decoder predicts individual instances by grouping corresponding pixels into clusters.
As there are already efficient architectures for semantic segmentation~\cite{SwiftNet-cvpr2019, esanet2021icra}, bottom-up approaches are often more efficient than top-down approaches~\cite{emsanet2022ijcnn}.
However, both approaches require an additional post-processing step for combining instance and semantic segmentation.
By contrast, end-to-end approaches, such as MaX-DeepLab~\cite{Max-Deeplab-cvpr2021}, directly predict the panoptic segmentation without additional post-processing.
While already achieving great performance, these methods are not established yet and also require complex architectures that currently do not focus on efficiency.

Instance orientation estimation can also be done in multiple ways.
For extracting the orientation, patch-based methods, such as \cite{Biternion-gcpr2015, DeepOrientation-iros2019, MTPersonPerception-iros2020}, can be used.
Another way for estimating the orientation of an instance is to estimate the pose of its 3D bounding box as shown in \cite{3DBBox-Estimation-cvpr2017, monocular3DObjectDetection-cvpr2019, ObjectsAreDifferent3D-cvpr2021}.
In contrast to the aforementioned approaches, EMSANet~\cite{emsanet2022ijcnn} follows the bottom-up idea and incorporates orientation estimation into the instance decoder in a dense-prediction manner.
This way, the computational overhead for orientation estimation is limited and averaging multiple predictions for an instance is enabled.

In this paper, we follow the bottom-up design of EMSANet for tackling the scene analysis tasks. 
However, to further speed up inference, we examine the lightweight MLP-based decoder of SegFormer.
\section{Efficient Multi-task Scene Analysis with RGB-D Transformers}
\label{sec:main}
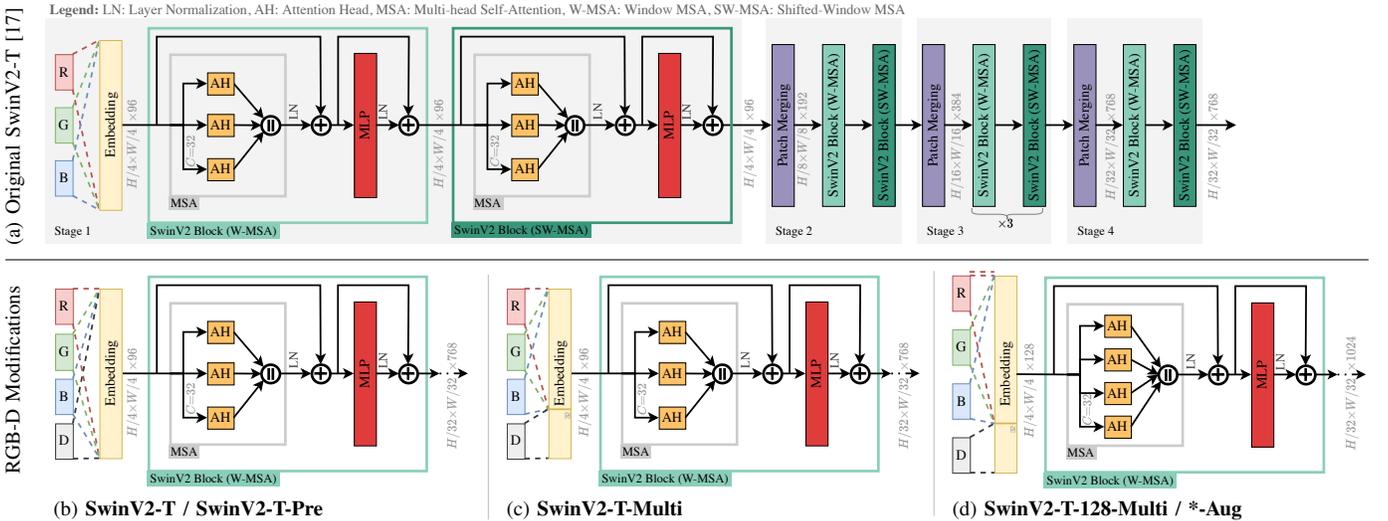
\begin{figure*}[!b]
    \centering%
    \vspace{-4mm}%
        \resizebox{\textwidth}{!}{%
            \begin{tikzpicture}[every node/.style={inner sep=0,outer sep=0}]%
                \node[anchor=north west] at (0.0, -0.05){\rotatebox{90}{\footnotesize (a) Original SwinV2-T~\cite{swinv2-cvpr2022}}};%
                \node[anchor=north west] at (0.5, 0){%
                    \includegraphics[scale=0.47, trim=0mm 0mm 0mm 5mm, clip]{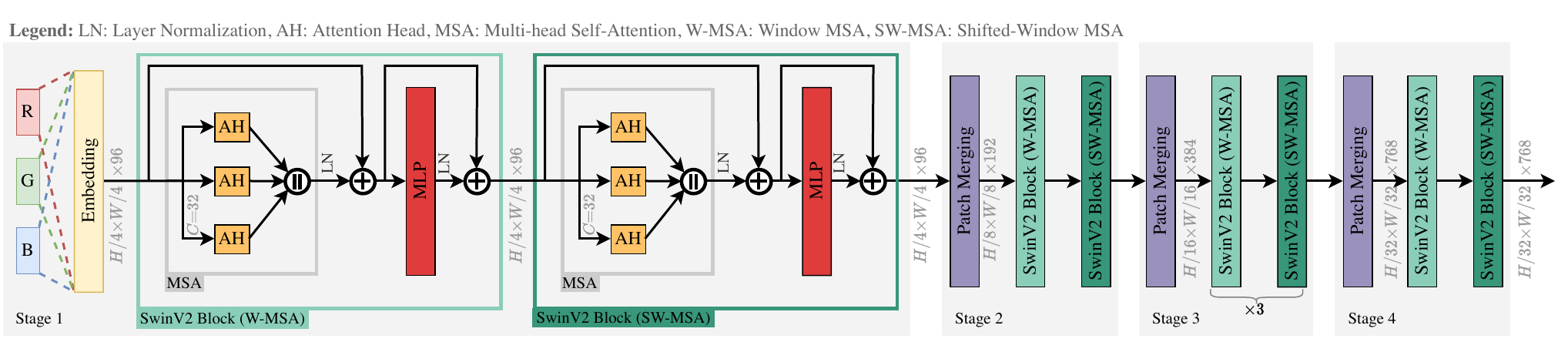}%
                };%
                \draw[thin] (0,-3.45) to (18.35,-3.45);
                \node[anchor=north west] at (0.42, -3.65){%
                    \includegraphics[scale=0.47, trim=0mm 0mm 0mm 0mm, clip]{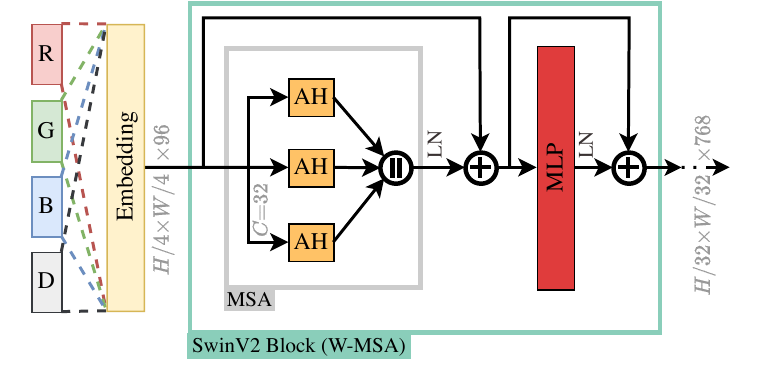}%
                };%
                \node[anchor=north west] at (6.5, -3.65){%
                    \includegraphics[scale=0.47, trim=0mm 0mm 0mm 0mm, clip]{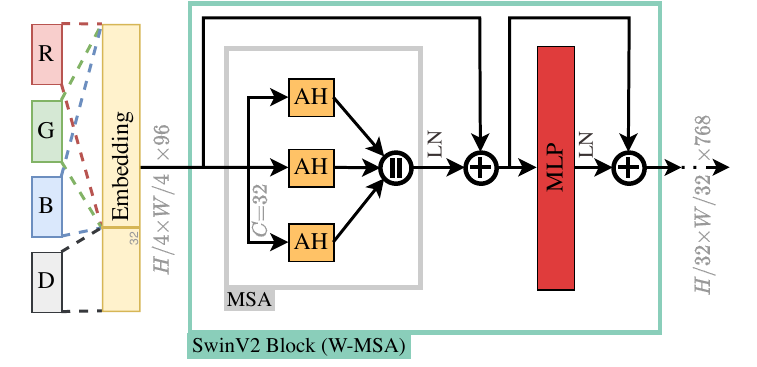}%
                };%
                
                \node[anchor=north west] at (12.5, -3.65){%
                    \includegraphics[scale=0.47, trim=0mm 0mm 4.5mm 0mm, clip]{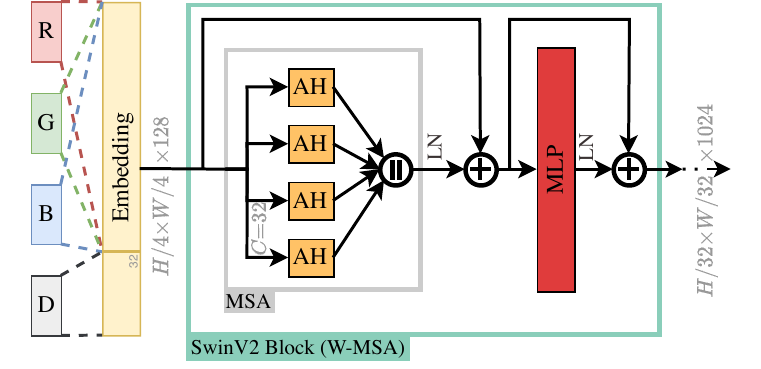}%
                };%
                \draw[thin, lightgray] (6.5,-3.65) -- (6.5,-6.95);
                \draw[thin, lightgray] (12.5,-3.6) -- (12.5,-6.9);
                \node[anchor=north west] at (0.0, -3.8){\rotatebox{90}{\footnotesize RGB-D Modifications}};%
                \node[anchor=north west] at (0.65, -6.7){\footnotesize (b) \textbf{SwinV2-T} / \textbf{SwinV2-T-Pre}};%
                \node[anchor=north west] at (6.75, -6.7){\footnotesize (c) \textbf{SwinV2-T-Multi}};%
                \node[anchor=north west] at (12.75, -6.7){\footnotesize (d) \textbf{SwinV2-T-128-Multi} / \textbf{*-Aug}};%
            \end{tikzpicture}%
        }%
    \vspace{-3mm}%
    \caption{%
        Original SwinV2-T architecture~\cite{swinv2-cvpr2022}~(top) and our modifications~(bottom) to efficiently incorporate depth information in a single encoder backbone. 
    }%
    \label{fig:swin}%
\end{figure*}

Our Transformer-based approach for scene analysis~(EMSAFormer) is shown in Fig.~\ref{fig:architecture}.
It builds on top of the EMSANet~\cite{emsanet2022ijcnn}.
Both architectures share a similar encoder-decoder design and the same task encoding. 
The encoder extracts semantically rich features from the input and performs downsampling up to a factor of \nicefrac{1}{32} of the input resolution to reduce computational effort.
However, instead of using a fused dual encoder, our EMSAFormer features only a single encoder with a modified SwinV2 Transformer backbone to incorporate RGB and depth data.
We address these modifications in Sec~\ref{sec:main:encoder}.
After the encoder, a context module~(CM) similar to the pyramid pooling module in PSPNet~\cite{PSPNet-cvpr2017} is attached. 
Although Transformers already enable a larger receptive field~\cite{SegFromer-neurips2021}, we observe a performance boost due to an additional context module in our experiments~(see results later in Fig.~\ref{fig:single_task}).
Due to the large downsampling at the end of the encoder, the computational effort of the context module is almost negligible.
While all tasks share the same encoder~(often referred to as hard-parameter sharing~\cite{MTL-survey-tpami-2021}), we use three independent decoders, not sharing any network parameters, to handle the tasks for scene analysis. 
We introduce the task-specific decoders in Sec~\ref{sec:main:decoders}.
Similar to EMSANet, the entire network architecture is tailored to enable fast inference.
However, as Transformer-based architectures are relatively new, inference optimization is crucial and currently rare. 
We address this key aspect with an additional NVIDIA TensorRT extension introduced in Sec~\ref{sec:main:optimization}.

\subsection{Encoder}
\label{sec:main:encoder}
The encoder of our EMSAFormer is derived from the SwinV2 Transformer~\cite{swinv2-cvpr2022} architecture.
We build on top of the tiny model SwinV2-T as it is the only version that currently enables real-time inference on our target hardware. 
The next larger SwinV2-S and SwinV2-B increase inference time by a factor of 1.5 and 1.9, respectively, and, thus, are out of our scope.
The architecture of SwinV2-T is shown Fig~\ref{fig:swin}~(a).
Each RGB input is processed in four stages. 
The first stage embeds the input using a $4{\times}4$ convolution with stride 4 to 96 feature maps. 
As this convolution processes non-overlapping patches of $4{\times}4$ pixels, this step is called patch embedding.
Subsequent to the patch embedding, the first two SwinV2 blocks are attached within the same stage. 
Each SwinV2 block comprises a multi-head self-attention module~(MSA) and a subsequent 2-layer multilayer perceptron~(MLP). 
Both the MSA and the MLP are followed by a layer normalization and further feature a skip connection as show in Fig.~\ref{fig:swin}.
The design of the SwinV2 block follows the original Transformer block introduced in~\cite{Transformer-neurips2017}.
However, in contrast to~\cite{Transformer-neurips2017, swin-iccv2021, SegFromer-neurips2021}, the attention is computed using a scaled cosine function instead of the softmax function.
Moreover, as computing the self-attention between all elements requires quadratic complexity, the authors adapt the MSA to a window multi-head self-attention (W-MSA) that divides the input in non-overlapping windows of size $8{\times}8$ in order to reduce complexity.
However, this approach lacks connections across windows and, thus, prevents the model from building context features.
To overcome this limitation, the authors further introduce a shifted-window multi-head self-attention~(SW-MSA).
By alternating both modules, the network is able to exchange information between adjacent windows.
For further details on the exact architecture, we refer to~\cite{swin-iccv2021, swinv2-cvpr2022}.
The remaining stages follow the same design.
However, the initial patch embedding is replaced with a patch merging operation, and the number of repeated SwinV2 blocks differs in stage 3.
The patch merging aims at reducing complexity while creating hierarchical features as the network gets deeper similar to CNN-based architectures~\cite{ResNet-cvpr2016, EfficientNet-icml2019, convnext-cvpr2022}.
To achieve this, the spatial resolution gets downsampled by a factor of 2, while the number of feature maps is doubled.

To incorporate depth data, we examine the modification depicted in Fig~\ref{fig:swin}~(bottom).
The most straight-forward way is shown in~Fig~\ref{fig:swin}~(b) and integrates depth as additional modality to the patch embedding.
The missing weights can be derived either by reusing the existing weights~(D=R+G+B) or by performing an additional pretraining step. 
We refer to both modifications as \emph{SwinV2-T} and \emph{SwinV2-T-Pre}, respectively.
Unfortunately, the ImageNet dataset~\cite{ImageNet-ijcv2015} used for pretraining does not feature depth data. 
Therefore, we use a grayscale image instead for pretraining~(D=gray). 
This way, the backbone already learns to handle four input channels during pretraining.

However, a major drawback of this approach is that both modalities are mixed right at the beginning of the network in the patch-embedding step.
As already shown in~\cite{FuseNet-accv2016}, such an early mixing of both modalities introduces issues due to deviating statistics and characteristics and eliminates any benefits.
Luckily, the Swin Transformer architecture enables to prevent mixing features up to the first patch merging.
Similar to most other Transformer-based architectures~\cite{Transformer-neurips2017, gpt-neurips2017, vit-iclr2021}, the attention is not computed across all channels but instead only on a subset of the channels.
In Swin Transformers, each attention head only processes a subset of 32 channels of the input to a SwinV2 block~(see MSA box in Fig~\ref{fig:swin}).
To take advantage of this property, only the patch embedding needs to be adapted. 
As shown in Fig~\ref{fig:swin}~(c), we propose to split the $4{\times}4$ convolution into two convolutions, the first embedding RGB to 64 channels and the second embedding depth to the remaining 32 channels.
We refer to this modification as \emph{SwinV2-T-Multi}.
Note that the MLPs subsequent to the multi-head self-attention blocks~(see Fig~\ref{fig:swin}~red) still combine channels. 
However, there is a skip connection, which means that the network is able to learn whether to combine features or not in an adjustable way.

While this approach focuses on independent processing of depth, both modalities are embedded to the 96 channels of original SwinV2-T model.
Embedding more information with the same number of channels, may lead to a bottleneck.
To overcome this issue, we further propose to enlarge the width of the entire model and to use 128, i.e., 96+32, channels in the initial embedding.
This way, the RGB embedding retains its original representation capabilities, and depth is embedded to additional 32 channels.
Note, due to the subset property of the attention heads, depth is still processed in independent attention heads.
We refer to this modification as \emph{SwinV2-T-128-Multi} in Fig~\ref{fig:swin}~(d).
Note that this modification leads to a width similar to the larger SwinV2-B.
However, the number of blocks and, thus, the depth remains the one of SwinV2-T.
To further strengthen the independent processing of depth in the subset of channels, we further add an additional augmentation step to the pretraining pipeline that randomly masks out either the whole RGB or grayscale image.
This way, the network is forced to learn to extract information from both images and cannot rely on the more meaningful RGB input solely.
We refer to this modification as \emph{SwinV2-T-128-Multi-Aug}.

In Sec~\ref{sec:experiments:single}, we examine the suitability of the aforementioned modifications.
We further investigated modification to the patch merging to extend our design principle, i.e., processing depth in an independent subset of channels with an adjustable fusion mechanism in the MLPs, to the entire architecture. 
However, we could not observe any significant performance improvement when adapting subsequent stages.

\subsection{Decoders}
\label{sec:main:decoders}
The decoders for our multi-task architecture~(see Fig~\ref{fig:architecture}) are designed to suit the specific needs of each task.
To obtain the final prediction for scene classification, a single fully-connected layer is attached to the global-average-pooling branch of the context module.
For panoptic segmentation, two dense decoders are used to perform semantic and instance segmentation.
Each decoder is followed by a task head that projects to the required number of channels for the corresponding tasks and, finally, restores the input resolution.
For semantic segmentation, the task head projects to the number of semantic classes.
For instance segmentation, we follow the bottom-up approach of Panoptic DeepLab~\cite{PanopticDeeplab-cvpr2020} and EMSANet~\cite{emsanet2022ijcnn}.
As shown in Fig~\ref{fig:architecture}, instances are represented by their center of mass encoded within a heatmap predicted by the first instance head.
To assign pixels to instance centers, a second instance head further predicts offset vectors pointing towards a corresponding instance center.
To ignore pixels belonging to stuff classes, e.g., wall or floor, a foreground mask derived from the semantic head is applied before assigning any pixel.
As shown in~\cite{emsanet2022ijcnn}, instance orientation estimation can also be handled in a dense manner with an additional head on top of the instance decoder.
For each pixel the orientation is predicted as continuous angle around the axis perpendicular to the ground plane.
To obtain the orientation of an instance, all predictions assigned to this instance are averaged.

For the dense decoders, we consider both the CNN-based decoder from EMSANet~(see red block in instance branch in Fig.~\ref{fig:architecture}) as well as the SegFormer MLP decoder~(see green block in semantic branch in Fig.~\ref{fig:architecture}).
The original SegFormer MLP decoder in~\cite{SegFromer-neurips2021} uses four branches with equal number of channels, i.e., $[128, 128, 128, 128]$ channels. 
We propose to follow the pyramid-like structure with increasing resolution of the EMSANet decoder and use $[256, 128, 64, 64]$ channels. 
This way, both inference throughput and performance are increased.
We refer to this decoder as modified SegFormer decoder.
In our experiments, we examine and compare the performance of both decoder types, the EMSANet decoder and the modified SegFormer decoder.

\subsection{Optimization}
\label{sec:main:optimization}
For efficient and fast inference on embedded hardware, optimized inference frameworks, such as NVIDIA TensorRT, are crucial.
Compared to CNN-based architectures, Transformer-based architectures are relatively new and, thus, currently lack the same kind of highly optimized inference engines. 
However, there is already ongoing effort to optimize inference throughput.
FasterTransformer~\cite{fastertransformer} provides a first extension to NVIDIA TensorRT with focus on Transformers.
However, while already archiving a significant speedup, they currently only focus on optimizing architectures for image classification with inputs of fixed and square resolution.
Moreover, the whole encoder is optimized as a single block, which makes it impossible to access intermediate features for skip connections or to incorporate any modification to the encoder architecture, such as a modified patch embedding or merging, a deviating number of channels, or another kind of normalization layer.
To overcome these limitations, we propose a custom NVIDIA TensorRT extension. 
It is based on the existing FasterTransformer implementation but splits the encoder in smaller blocks to enable more flexibility in downstream tasks. 
We further added support for bottom-right padding to the CUDA kernels if the input size is not a multiple of the window size used in the shifted-window attention. 
This is of great importance to enable inference with inputs of arbitrary input resolution.
The modular design is complemented by the ability to import models from Open Neural Network Exchange (ONNX) format.
The proposed extension enables us to examine the architecture modifications described in Sec~\ref{sec:main:encoder}.
\section{Experiments}
\label{sec:experiments}
We evaluate the performance of our proposed multi-task approach on the common indoor RGB-D datasets NYUv2~\cite{NYUv2-eccv2012}, SUNRGB-D~\cite{SUNRGBD-cvpr2015}, and ScanNet~\cite{scannet-cvpr2017}.
We start with a single-task setting on the smaller NYUv2 dataset to assess the performance of the SwinV2-based encoder and our proposed modifications across the individual tasks. 
The goal is to derive a suitable encoder that is capable of handling all tasks in a most efficient way.
Moreover, we investigate the performance of our proposed encoder with both dense decoder types, the EMSANet decoder and the modified SegFormer decoder. 
With the results of these experiments at hand, we combine all tasks in a multi-task approach.
The goal is to solve all four tasks, i.e., semantic segmentation, instance segmentation, instance orientation estimation, and scene classification, simultaneously using a single neural network.
Finally, we demonstrate the suitability of our approach for the larger SUNRGB-D and ScanNet dataset and compare to other state-of-the-art approaches.

\subsection{Implementation Details}
\label{sec:experiments:implementation}
Our implementation is built using PyTorch~\cite{pytorch-neurips2019} and is based on the EMSANet implementation~\cite{emsanet2022ijcnn}.
As commonly done in downstream tasks, we use weights pretrained on ImageNet~\cite{ImageNet-ijcv2015} to initialize the encoders.
However, as already stated in Sec.~\ref{sec:main:encoder}, any modification except for SwinV2-T~(D=R+G+B) requires an additional pretraining step. 
Pretraining Transformers from scratch is very time consuming and requires large batch sizes and, thus, heavy memory requirements. 
We used $8\times$ NVIDIA A100 40$\,$GB GPUs to accomplish these pretrainings.
To enable other applications, we share the pretrained weights to the research community on GitHub.
Note that pretraining larger models, such as SwinV2-S or SwinV2-B, implies even higher memory requirements and, thus, is out of our scope.
Subsequent training of the EMSAFormer architecture requires less resources and can be done on any GPU with at least 25$\,$GB of VRAM (all tasks simultaneously).
We stick to the training pipeline of EMSANet~\cite{emsanet2022ijcnn} for data processing and augmentation. 
We use a fixed input resolution of $640\times480$ pixels and a batch size of 8.
Each network is trained for 500 epochs using SGD with a momentum of 0.9 and a small weight decay of 0.0001.
The learning rate is varied in $\{0.00125, 0.0025, 0.005, 0.01, 0.02, 0.03, 0.04, 0.06\}$ depending on the actual dataset and tasks.
We further use a one-cycle learning rate scheduler, similar to the for Transformer commonly used cosine-annealing learning rate scheduler, to adjust the learning during training.
For further details and other hyperparameters, we refer to our implementation available on GitHub.
\begin{figure*}[!b]
	\centering%
	\vspace{-4.5mm}%
	\resizebox{\textwidth}{!}{%
	\begin{tikzpicture}[scale=0.99]%
		\node[anchor=north, rotate=90] at (0, -2.5){\footnotesize{(a) Semantic Segmentation (Sem)}};%
	    \node[anchor=north west] at (0.5, 0){%
	        \includegraphics[scale=0.59]{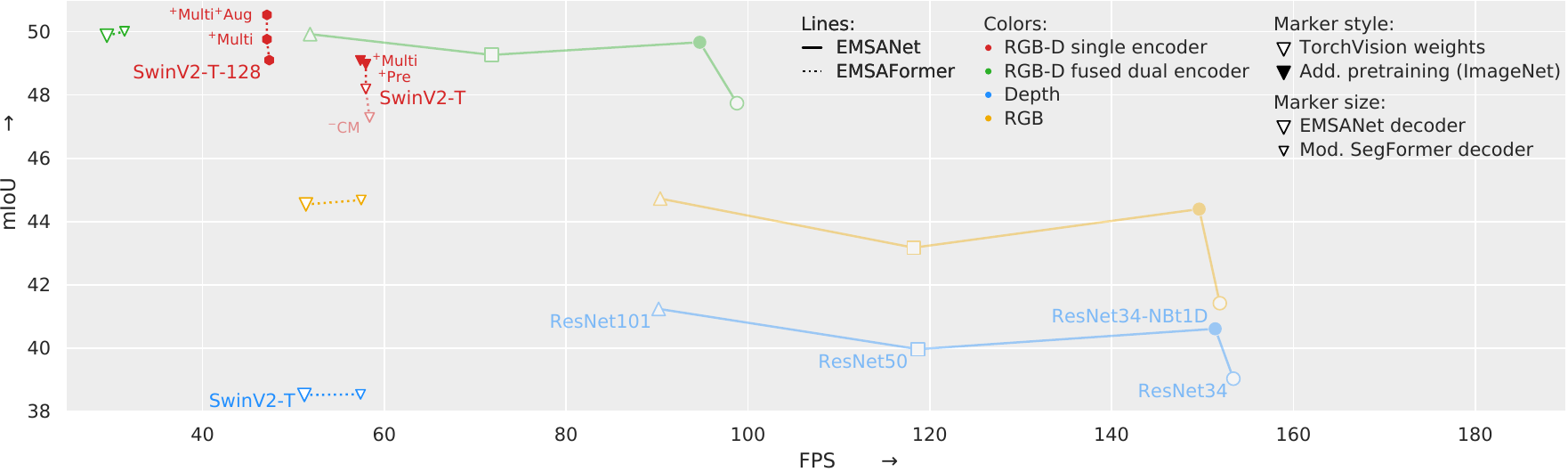}%
	    };
		\node[anchor=north, rotate=90] at (0, -8.05){\footnotesize{(b) Instance Segmentation (Ins)}};%
	    \node[anchor=north west] at (0.5, -5.55){%
	        \includegraphics[scale=0.59]{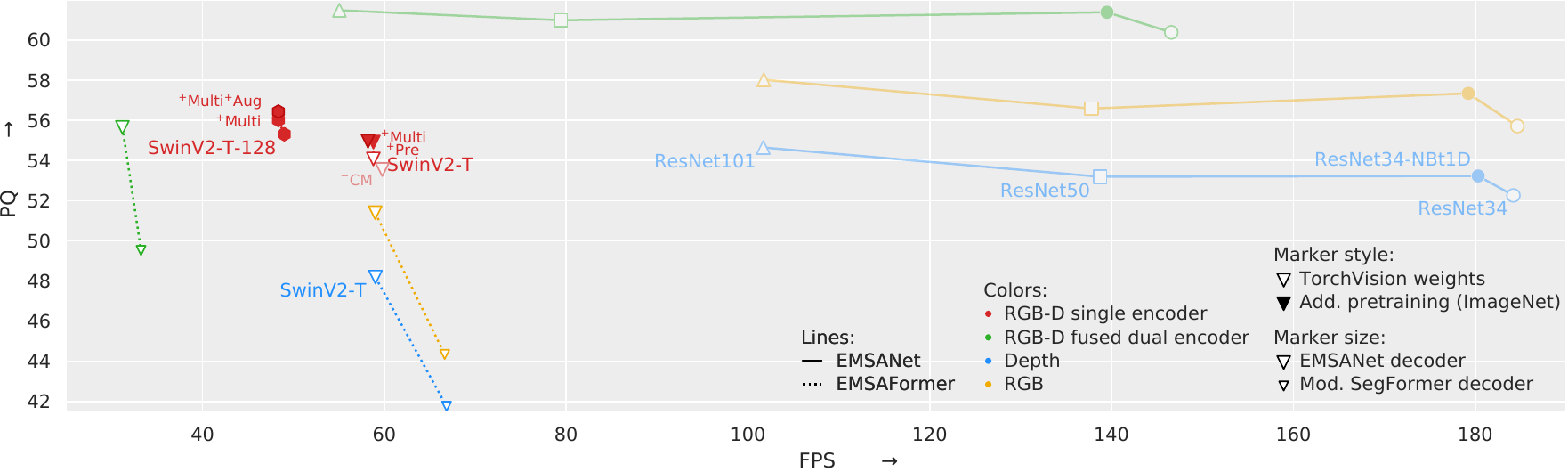}%
	    };	    
	\end{tikzpicture}%
    }
	\vspace{-8.5mm}
	\caption{%
	Results on NYUv2 test split when performing semantic segmentation~(top) and instance segmentation~(bottom) in a single-task setting with various encoder configurations over inference throughput~(NIVIDA Jetson AGX Orin 32$\,$GB, Jetpack 5.1.1, TensorRT 8.5.2, Float16, 50$\,$W). %
	See Sec.~\ref{sec:experiments:metrics} for metrics.%
	}
	\label{fig:single_task}
\end{figure*}

\subsection{Datasets}
\label{sec:experiments:datasets}
Selecting datasets suitable for evaluating our multi-task approach is challenging as it requires RGB and depth data as well as annotations for the individual tasks.
In the following, we give a brief overview over the RGB-D datasets used.

\textit{NYUv2~\cite{NYUv2-eccv2012}:} \enspace %
The NYUv2 dataset comprises 795 training samples and 654 test samples. 
It provides annotations for semantic and instance segmentation and scene classification.
We use the semantic annotations with 40 classes.
As the original dataset does not include annotations for instance orientation, we use the manually annotated ones from~\cite{emsanet2022ijcnn}.

\textit{SUNRGB-D~\cite{SUNRGBD-cvpr2015}:} \enspace %
The SUNRGB-D dataset features 5,285 training and 5,050 test samples from multiple RGB-D cameras.
The dataset provides annotations for the first 37~NYUv2 semantic classes and for scene classification. 
However, it lacks dense annotations for instance segmentation.
We stick to the reconstructed instance annotations from 3D bounding boxes proposed in ~\cite{emsanet2022ijcnn}, which also provide orientation annotations.

\textit{ScanNet~\cite{scannet-cvpr2017}:} \enspace %
The ScanNet dataset comprises 1.89M training, 0.53M validation, and 0.21M test images.
It provides annotations for semantic and instance segmentation as well as for scene classification. 
We use the semantic class mapping to the 40~NYUv2 classes.
As the dataset is created from video sequences and, thus, contains many similar images, we follow the official recommendation~\cite{scannet-cvpr2017} and use the subsample of 100 for the validation and test split.
To reduce training time, we use a subsample of 50 and limit the number of samples to a random subset of 25\% for each epoch.
As the dataset lacks instance orientation annotations, we cannot train this task on ScanNet.
However, given the size and the quality of the annotations, it is still an important dataset.

For creating panoptic annotations, we combine the dense annotations of the datasets and treat floor, wall, and ceiling as stuff classes.
For scene classification, we use the unified class spectrum for the most relevant indoor classes presented in~\cite{emsanet2022ijcnn}. 
As the ScanNet dataset was not in the scope of~\cite{emsanet2022ijcnn}, we created a similar scene class mapping for this dataset.

\subsection{Metrics}
\label{sec:experiments:metrics}
We follow the evaluation protocol of~\cite{emsanet2022ijcnn} and report the mean intersection over union~(mIoU) for semantic segmentation, panoptic quality~(PQ), segmentation quality~(SQ) and recognition quality~(RQ) for panoptic segmentation as well as the mean absolute angular error~(MAAE) for instance orientation estimation.
To enable experiments in a single-task setting, PQ is also reported for instance segmentation.
Note that PQ tracks closely to the average precision~(AP)~\cite{Panopticsegmentation-cvpr2019} and, thus, also evaluates instance segmentation in a meaningful way.
However, the instance decoder performs class-agnostic instance segmentation.
Therefore, we use the ground-truth semantic segmentation for creating a foreground mask and for assigning semantic classes.
The reported PQ is equal to perfect semantic segmentation.
Similarly, for single-task instance orientation estimation, the MAAE is computed assuming perfect instances.
For scene classification, the balanced accuracy~(bAcc) is used to account for the imbalanced class distribution.
As we aim at fast inference, we do not apply any evaluation tricks, such as test time augmentation.
Furthermore, to enable fair comparison, we always upsample dense predictions to the full input resolution before determining any metric.
\subsection{Single-task Performance}
\label{sec:experiments:single}
We start by evaluating the proposed SwinV2-Transformer-based encoder and its modifications in a single-task setting.

\setcounter{table}{1}
\begin{table*}[!b]
\vspace{-5mm}
\centering
\caption{%
    Results on NYUv2 test split when training our multi-task EMSAFormer and in comparison to EMSANet~\cite{emsanet2022ijcnn}. %
    See Sec.~\ref{sec:experiments:metrics} for the reported metrics. %
    Panoptic results are obtained after merging semantic and instance prediction.
    Legend: italic:~metric used for determining the best checkpoint, gray:~best result within the same run, FPS$_{\text{x}}$:~frames per second on an NVIDIA Jetson AGX Orin 32$\,$GB (Jetpack 5.1.1, TensorRT 8.5.2, Float16) at measured power consumption $\text{x}$.
}
\vspace{-2.5mm}
\scalebox{0.91}{%
\scriptsize%
\begin{tabular}{@{\hspace{1mm}}c@{\hspace{2.5mm}}ll@{\hspace{5mm}}c@{\hspace{7mm}}c@{\hspace{2mm}}c@{\hspace{7mm}}c@{\hspace{7mm}}c@{\hspace{2mm}}c@{\hspace{2mm}}c@{\hspace{2mm}}c@{\hspace{1mm}}c@{\hspace{7mm}}c@{\hspace{2mm}}c@{\hspace{1mm}}}
\toprule
& & & \textbf{\begin{tabular}[c]{@{}c@{}}\color[HTML]{737373}Semantic\\ \color[HTML]{737373}Decoder\end{tabular}} & \multicolumn{2}{c@{\hspace{9mm}}}{\textbf{\begin{tabular}[c]{@{\hspace{-2.5mm}}c@{}}\color[HTML]{737373}Instance\\ \color[HTML]{737373}Decoder\end{tabular}}} & \textbf{\begin{tabular}[c]{@{}c@{}}\color[HTML]{5c5c5c}Scene\\ \color[HTML]{5c5c5c}Head\end{tabular}} & \multicolumn{5}{c@{\hspace{9mm}}}{\textbf{\begin{tabular}[c]{@{}c@{}}\color[HTML]{737373}Panoptic Results\\ \color[HTML]{737373}(after merging)\end{tabular}}} & \multicolumn{2}{c@{\hspace{4mm}}}{\textbf{\begin{tabular}[c]{@{}c@{}}\color[HTML]{737373}Inference\\ \color[HTML]{737373}Throughput\end{tabular}}} \\
& \textbf{Backbone} & \textbf{Task(s)} & \textbf{mIoU}$\,{}^\uparrow$ & \textbf{PQ}$\,{}^\uparrow$ & \textbf{MAAE}$\,{}^\downarrow$ & \textbf{bAcc}$\,{}^\uparrow$ & \textbf{mIoU}$\,{}^\uparrow$ & \textbf{PQ}$\,{}^\uparrow$ & \textbf{RQ}$\,{}^\uparrow$ & \textbf{SQ}$\,{}^\uparrow$ & \textbf{MAEE}$\,{}^\downarrow$ & \textbf{FPS}$\,{}^\uparrow_{\text{50W}}$ & \textbf{FPS}$\,{}^\uparrow_{\text{30W}}$ \\ \midrule
\multirow{10}{*}{\rotatebox{90}{\textbf{\color[HTML]{9B9B9B}~~~EMSAFormer (ours)}}} & \textbf{SwinV2-T-128-Multi-Aug} & Semantic Segmentation (Sem) & %
    50.53 & --- & --- & --- & --- & --- & --- & --- & --- & 47.1 & 30.5 \\%
&  & Instance Segmentation (Ins) & %
    --- & 56.44 & --- & --- & --- & --- & --- & --- & --- & 48.4 & 33.5 \\%
&  & Orientation Estimation (Or) & %
    --- & --- & 17.85 & --- & --- & --- & --- & --- & --- & 47.7 & 32.9 \\%
&  & Scene Classification (Sce) & %
    --- & --- & --- & 78.66 & --- & --- & --- & --- & --- & 58.9 & 40.7 \\%
\cmidrule{3-14}
&  & Sem(SegFormer)$\,$+$\,$Sce$\,$+$\,$Ins$\,$+$\,$Or & %
    50.23 & 58.75 & 20.95 & 77.70 & 51.34 & \emph{43.41} & 52.53 & 81.75 & 18.94 & 39.1 & 27.3 \\
& &  & %
    {\color[HTML]{9B9B9B} 50.51} & {\color[HTML]{9B9B9B} 59.25} & {\color[HTML]{9B9B9B} 20.95} & {\color[HTML]{9B9B9B} 80.02} & {\color[HTML]{9B9B9B} 51.34} & {\color[HTML]{9B9B9B} \emph{43.41}} & {\color[HTML]{9B9B9B} 52.53} & {\color[HTML]{9B9B9B} 81.79} & {\color[HTML]{9B9B9B} 18.94} & & \\%
\cmidrule{3-14}
&  & Sem$\,$+$\,$Sce$\,$+$\,$Ins$\,$+$\,$Or & %
    51.06 & 59.06 & 20.01 & 78.80 & 51.76 & \emph{43.28} & 52.48 & 81.43 & 18.26 & 36.5 & 25.6 \\
& &  & %
    {\color[HTML]{9B9B9B} 51.26} & {\color[HTML]{9B9B9B} 59.27} & {\color[HTML]{9B9B9B} 18.09} & {\color[HTML]{9B9B9B} 78.80} & {\color[HTML]{9B9B9B} 51.76} & {\color[HTML]{9B9B9B} \emph{43.28}} & {\color[HTML]{9B9B9B} 52.48} & {\color[HTML]{9B9B9B} 81.51} & {\color[HTML]{9B9B9B} 18.09} & & \\%
\midrule\midrule
\multirow{5}{*}{\rotatebox{90}{\textbf{\color[HTML]{9B9B9B}~~EMSANet}}} & \textbf{2x ResNet101} & Sem$\,$+$\,$Sce$\,$+$\,$Ins$\,$+$\,$Or & %
    50.83 & 62.64 & 17.87 & 77.41 & 50.67 & \emph{45.12} & 54.02 & 82.49 & 15.33 & 42.9 & 30.1 \\
& &  & %
    {\color[HTML]{9B9B9B} 51.01} & {\color[HTML]{9B9B9B} 62.81} & {\color[HTML]{9B9B9B} 17.82} & {\color[HTML]{9B9B9B} 78.43} & {\color[HTML]{9B9B9B} 51.23} & {\color[HTML]{9B9B9B} \emph{45.12}} & {\color[HTML]{9B9B9B} 54.02} & {\color[HTML]{9B9B9B} 82.99} & {\color[HTML]{9B9B9B} 14.73} & & \\%
\cmidrule{2-14}
& \textbf{2x ResNet34-NBt1D} & Sem$\,$+$\,$Sce$\,$+$\,$Ins$\,$+$\,$Or & %
    50.97 & 61.33 & 18.37 & 76.46 & 50.61 & \emph{43.59} & 52.23 & 82.48 & 16.39 & 70.5 & 49.9 \\
& &  & %
    {\color[HTML]{9B9B9B} 51.15} & {\color[HTML]{9B9B9B} 61.53} & {\color[HTML]{9B9B9B} 18.37} & {\color[HTML]{9B9B9B} 78.18} & {\color[HTML]{9B9B9B} 51.31} & {\color[HTML]{9B9B9B} \emph{43.59}} & {\color[HTML]{9B9B9B} 52.27} & {\color[HTML]{9B9B9B} 82.70} & {\color[HTML]{9B9B9B} 15.76} & & \\%
\bottomrule
\end{tabular}
}
\label{tab:results_nyuv2}
\vspace{-1mm}
\end{table*}
\setcounter{table}{0}

\textit{Semantic Segmentation~(Sem):} \enspace %
Fig~\ref{fig:single_task}~(a) visualizes the results for semantic segmentation and compares to the dual-encoder approaches of EMSANet.
It becomes obvious that SwinV2-T can also be used as backbone in a dual-encoder design leading to an mIoU similar to one with ResNet101 backbones except for processing depth solely~(blue in Fig~\ref{fig:single_task}).
This highlights that SwinV2 is tailored to processing RGB inputs.
However, the dual-encoder design results in a significant drop in inference throughput, making such a design not suitable for our application scenario.
Changing the decoder to the smaller modified SegFormer decoder leads to similar performance but cannot alleviate the gap in inference throughput.
By contrast, relying on a single encoder greatly improves inference throughput.
However, the results~(red in Fig~\ref{fig:single_task}) also highlight the challenge of processing both modalities in a single encoder.
The performance of SwinV2-T drops to an mIoU of \textasciitilde48\%. 
Additional pretraining on RGB-Gray inputs~(SwinV2-T-Pre) can only halve the gap to the dual-encoder counterpart~(green in Fig~\ref{fig:single_task}).
Further splitting the patch embedding to emphasize processing both modalities independently, as done in SwinV2-T-Multi, does not close the remaining gap in performance.
This shows that the model is not capable of handling both modalities in the original embedding with 96 channels.
The wider SwinV2-T-128, which uses 128 instead of 96 channels in the patch embedding, benefits much more from splitting the patch embedding~(SwinV2-T-128-Multi).
Adapting data augmentation during pretraining~(SwinV2-T-128-Multi-Aug) to further strengthen independent processing of depth later in the downstream tasks results in another improvement.
Our single encoder with SwinV2-T-128-Multi-Aug backbone leads to slightly better performance than the dual-encoder design with ResNet101 backbone at almost the same inference speed.

\textit{Instance Segmentation~(Ins):} \enspace %
For instance segmentation, a similar trend is emerged. 
However, the results in Fig~\ref{fig:single_task}~(b) show two new aspects.
First, there is a gap of \textasciitilde6\% in PQ between CNN-based and Transformer-based encoders independently of the modality or the encoder design.
As discussed later in Sec.~\ref{sec:experiments:multi}, this gap highlights optimization issues in the Transformer-based encoder due to a small number of training samples along with a more challenging task.
Second, the modified SegFormer MLP decoder leads consistently to even worse results.
Therefore, we stick to the EMSANet decoder for instance segmentation for the remaining experiments.

\begin{table}[t!]
\scriptsize
\centering
\caption{%
    Results on NYUv2 test split when performing instance orientation estimation~(top) and scene classification~(bottom) in a single-task setting with various encoder configurations. %
    See Sec.~\ref{sec:experiments:metrics} for metric abbreviations.%
}
\vspace{-2.5mm}
\begin{tabular}{@{\hspace{1mm}}l@{\hspace{2.5mm}}l@{\hspace{5mm}}c@{\hspace{2.5mm}}c@{\hspace{5mm}}cc@{\hspace{1mm}}}%
\toprule%
\multicolumn{2}{@{\hspace{1mm}}l}{\textbf{Orientation Estimation}}                       &  & \multicolumn{1}{l}{} & \multicolumn{2}{c}{\textbf{RGB-D}} \\%
\multicolumn{2}{@{\hspace{1mm}}l}{\textbf{MAAE}$\,{}^\downarrow$}                        & \textbf{RGB}    & \textbf{Depth}    & \textbf{Fused Dual}   & \textbf{Single}    \\%
\midrule%
\multirow{3}{*}{\rotatebox{90}{\textbf{\color[HTML]{9B9B9B}\cite{emsanet2022ijcnn}~$\;$}}}
                            & ResNet34-NBt1d                                  & 22.24           & 18.36             & 17.91                 & ---            \\%
                            & ResNet50                                        & 23.09           & 18.81             & 18.41                 & ---            \\%
                            & ResNet101                                       & 22.06           & 18.02             & \textbf{17.50}        & ---            \\%
\midrule%
                            & SwinV2-T                                   & 24.08           & 19.13             & 19.02                 & 19.52          \\%
\midrule%
\multirow{3}{*}{\rotatebox{90}{\color[HTML]{9B9B9B}\textbf{Ours}}} & SwinV2-T-Pre                     & ---             & ---               & ---                   & 18.91          \\%
                            & SwinV2-T-128                                    & ---             & ---               & ---                   & 18.89          \\%
                            & SwinV2-T-128-Multi-Aug                          & ---             & ---               & ---                   & \textbf{17.85} \\%
\bottomrule                            
\end{tabular}
\label{tab:results_nyuv2_st}

\vspace{2mm}
\begin{tabular}{@{\hspace{1mm}}l@{\hspace{2.5mm}}l@{\hspace{5mm}}c@{\hspace{2.5mm}}c@{\hspace{5mm}}cc@{\hspace{1mm}}}%
\toprule%
\multicolumn{2}{@{\hspace{1mm}}l}{\textbf{Scene Classification}}                       &  & \multicolumn{1}{l}{} & \multicolumn{2}{c}{\textbf{RGB-D}} \\%
\multicolumn{2}{@{\hspace{1mm}}l}{\textbf{bAcc}$\,{}^\uparrow$}                   & \textbf{RGB}    & \textbf{Depth}    & \textbf{Fused Dual}   & \textbf{Single}    \\%
\midrule%
\multirow{3}{*}{\rotatebox{90}{\textbf{\color[HTML]{9B9B9B}\cite{emsanet2022ijcnn}~$\;$}}}
                            & ResNet34-NBt1d                                  & 74.40           & 67.26             & 72.40                 & ---            \\%
                            & ResNet50                                        & 74.19           & 69.92             & 74.91                 & ---            \\%
                            & ResNet101                                       & 74.95           & 70.53             & \textbf{75.86}        & ---            \\%
\midrule%
                            & SwinV2-T                                   & 76.84           & 66.72             & 73.39                 & 76.52            \\%
\midrule%
\multirow{3}{*}{\rotatebox{90}{\color[HTML]{9B9B9B}\textbf{Ours}}} & SwinV2-T-Pre                     & ---             & ---               & ---                   & 77.32          \\%
                            & SwinV2-T-128                                    & ---             & ---               & ---                   & 78.21          \\%
                            & SwinV2-T-128-Multi-Aug                          & ---             & ---               & ---                   & \textbf{78.66} \\%
\bottomrule%
\end{tabular}%
\vspace{-4mm}%
\end{table}
\setcounter{table}{2}

\textit{Orientation Estimation~(Or):} \enspace %
The results in Tab.~\ref{tab:results_nyuv2_st}~(top) show that the gap experienced for instance segmentation is not present for instance orientation estimation.
This could be due to the fact that this task is easier to accomplish in general and of more similar complexity to semantic segmentation.
Our SwinV2-T-128-Multi-Aug encoder achieves comparable performance to a dual encoder with ResNet101 backbone, while outperforming all other dual-encoder approaches.

\textit{Scene Classification~(Sce):} \enspace %
The results in Tab.~\ref{tab:results_nyuv2_st}~(bottom) again highlight the strength of Transformer-based 
architectures for image classification.
The performance of the Transformer-based single encoder model processing RGB solely already outperforms all CNN-based approaches with ResNet backbone.
Furthermore, each single RGB-D encoder model outperforms all dual-encoder approaches.
Our proposed SwinV2-T-128-Multi-Aug achieves the best result.

The results of this set of experiments show that our proposed SwinV2-T-128-Multi-Aug backbone is capable of handling all tasks.
Issues for instance segmentation are addressed below.

\subsection{Multi-task Performance}
\label{sec:experiments:multi}
Learning multiple tasks using a single neural network is challenging as the tasks may influence each other. 
Thus, balancing the losses to each other and selecting the best epoch are crucial.
We put less focus on orientation estimation as the results are already close to annotation quality~\cite{emsanet2022ijcnn}.
However, as we focus on the Transformer-based encoder in this publication, we refer to our implementation for the actual task balancing.
The best epoch is chosen based on the task most relevant for our application, i.e., the PQ for panoptic segmentation.
However, to get a better impression on the performance of the individual tasks when trained simultaneously in a multi-task setting and independent of selecting a specific checkpoint, we also report the best result for each metric within the same run.

Tab~\ref{tab:results_nyuv2} shows the multi-task results for NYUv2.
It becomes obvious that all tasks can be solved using a single neural network. 
The results for the individual tasks are close to single-task performance.
The PQ for instance segmentation increases noticeably and closes the gap to CNN-based dual encoders partially. 
This indicates that the encoder features learned in the multi-task setting are more beneficial for instance segmentation. 
Surprisingly, exchanging the modified SegFormer decoder~(denoted by Sem(SegFormer) in Tab.~\ref{tab:results_nyuv2}) with the EMSANet decoder~(denoted by Sem in Tab.~\ref{tab:results_nyuv2}) for semantic segmentation improves almost the entire multi-task performance.
This suggests that -- at least for the smaller NYUv2 dataset -- two identical dense decoders lead to better encoder features.
Except for instance segmentation, the multi-task results are close to EMSANet with ResNet101 backbone.

The results for SUNRGB-D and ScanNet in Tab.~\ref{tab:results_sunrgbd_scannet2} show a different picture.
Performing semantic segmentation with the modified SegFormer decoder (denoted by Sem(SegFormer) in Tab.~\ref{tab:results_sunrgbd_scannet2}) consistently leads to better multi-task results.
The gap to the EMSANet decoder further gets larger as the dataset size increases, i.e., for ScanNet.
We deduce that the NYUv2 dataset is too small to fit the parameters of our model with modified SegFormer decoder.
The proposed EMSAFormer configuration~(see Fig~\ref{fig:architecture}) outperforms both EMSANet approaches for semantic and panoptic segmentation and scene classification.

Tab~\ref{tab:results_nyuv2} further reports the inference throughput for all approaches and task settings.
Note that the values also apply for SUNRGD-D and ScanNet as they feature less or the same number of semantic classes.
\begin{table*}[t!]

\centering
\caption{%
    Results on SUNRGB-D test split and ScanNet validation split when training our multi-task EMSAFormer and in comparison to EMSANet~\cite{emsanet2022ijcnn}. %
    See Sec.~\ref{sec:experiments:metrics} for details on the reported metrics. %
    Legend: italic:~metric used for determining the best checkpoint, gray:~best result within the same run.
}
\vspace{-2.5mm}
\scalebox{0.91}{%
\scriptsize%
\begin{tabular}{@{\hspace{1mm}}l@{\hspace{5mm}}ll@{\hspace{5mm}}c@{\hspace{7mm}}c@{\hspace{2mm}}c@{\hspace{7mm}}c@{\hspace{7mm}}c@{\hspace{2mm}}c@{\hspace{2mm}}c@{\hspace{2mm}}c@{\hspace{1mm}}c@{\hspace{1mm}}}
\toprule
& & & \textbf{\begin{tabular}[c]{@{}c@{}}\color[HTML]{737373}Semantic\\ \color[HTML]{737373}Decoder\end{tabular}} & \multicolumn{2}{c@{\hspace{9mm}}}{\textbf{\begin{tabular}[c]{@{\hspace{-2.5mm}}c@{}}\color[HTML]{737373}Instance\\ \color[HTML]{737373}Decoder\end{tabular}}} & \textbf{\begin{tabular}[c]{@{}c@{}}\color[HTML]{5c5c5c}Scene\\ \color[HTML]{5c5c5c}Head\end{tabular}} & \multicolumn{5}{c@{\hspace{9mm}}}{\textbf{\begin{tabular}[c]{@{\hspace{4.5mm}}c@{}}\color[HTML]{737373}Panoptic Results\\ \color[HTML]{737373}(after merging)\end{tabular}}} \\
& & \textbf{Model} & \textbf{mIoU}$\,{}^\uparrow$ & \textbf{PQ}$\,{}^\uparrow$ & \textbf{MAAE}$\,{}^\downarrow$ & \textbf{bAcc}$\,{}^\uparrow$ & \textbf{mIoU}$\,{}^\uparrow$ & \textbf{PQ}$\,{}^\uparrow$ & \textbf{RQ}$\,{}^\uparrow$ & \textbf{SQ}$\,{}^\uparrow$ & \textbf{MAEE}$\,{}^\downarrow$ \\ \midrule
\textbf{SUNRGB-D} & \textbf{EMSAFormer} & \textbf{SwinV2-T-128-Multi-Aug} & %
    48.52 & 61.14 & 16.99 & 62.01 & 45.12 & \emph{50.08} & 59.08 & 84.68 & 15.32 \\
& &  & %
    {\color[HTML]{9B9B9B} 48.67} & {\color[HTML]{9B9B9B} 61.60} & {\color[HTML]{9B9B9B} 16.99} & {\color[HTML]{9B9B9B} 64.96} & {\color[HTML]{9B9B9B} 45.27} & {\color[HTML]{9B9B9B} \emph{50.08}} & {\color[HTML]{9B9B9B} 59.08} & {\color[HTML]{9B9B9B} 84.83} & {\color[HTML]{9B9B9B} 14.93} \\%
\cmidrule{3-12}
& & \textbf{SwinV2-T-128-Multi-Aug} (Sem(SegFormer)) & %
    48.61 & 61.20 & 15.91 & 61.97 & 45.79 & \emph{51.70} & 60.12 & 84.65 & 14.00 \\
& &  & %
    {\color[HTML]{999999} 48.82} & {\color[HTML]{999999} 61.78} & {\color[HTML]{999999} 15.91} & {\color[HTML]{999999} 64.50} & {\color[HTML]{999999} 45.94} & {\color[HTML]{999999} \emph{51.70}} & {\color[HTML]{999999} 60.15} & {\color[HTML]{999999} 84.65} & {\color[HTML]{999999} 13.90} \\%
\cmidrule{2-12}
& \textbf{EMSANet} & \textbf{2x ResNet101} & %
    \emph{47.99} & 62.07 & 15.17 & 59.40 & 43.22 & 51.06 & 58.88 & 85.53 & 13.34 \\
& &  & %
    {\color[HTML]{9B9B9B} \emph{47.99}} & {\color[HTML]{9B9B9B} 62.96} & {\color[HTML]{9B9B9B} 15.17} & {\color[HTML]{9B9B9B} 61.21} & {\color[HTML]{9B9B9B} 44.19} & {\color[HTML]{9B9B9B} 51.75} & {\color[HTML]{9B9B9B} 59.74} & {\color[HTML]{9B9B9B} 85.64} & {\color[HTML]{9B9B9B} 13.00} \\%
\cmidrule{3-12}
& & \textbf{2x ResNet34-NBt1D} & %
    \emph{48.39} & 60.62 & 16.28 & 61.76 & 45.53 & 49.88 & 57.79 & 84.91 & 14.23 \\
& &  & %
    {\color[HTML]{9B9B9B} \emph{48.39}} & {\color[HTML]{9B9B9B} 61.48} & {\color[HTML]{9B9B9B} 14.83} & {\color[HTML]{9B9B9B} 62.66} & {\color[HTML]{9B9B9B} 45.66} & {\color[HTML]{9B9B9B} 50.53} & {\color[HTML]{9B9B9B} 58.66} & {\color[HTML]{9B9B9B} 85.20} & {\color[HTML]{9B9B9B} 14.15} \\%
\midrule\midrule
\textbf{ScanNet} & \textbf{EMSAFormer} & \textbf{SwinV2-T-128-Multi-Aug} & %
    63.78 & 66.69 & --- & 48.82 & 61.93 & \emph{49.70} & 59.15 & 83.31 & --- \\
& &  & %
    {\color[HTML]{9B9B9B} 63.78} & {\color[HTML]{9B9B9B} 66.71} & {\color[HTML]{9B9B9B} ---} & {\color[HTML]{9B9B9B} 49.70} & {\color[HTML]{9B9B9B} 61.93} & {\color[HTML]{9B9B9B} \emph{49.70}} & {\color[HTML]{9B9B9B} 59.15} & {\color[HTML]{9B9B9B} 83.36} & {\color[HTML]{9B9B9B} ---} \\%
\cmidrule{3-12}
& & \textbf{SwinV2-T-128-Multi-Aug} (Sem(SegFormer)) & %
    64.75 & 67.71 & --- & 49.69 & 62.66 & \emph{51.18} & 61.01 & 83.20 & --- \\
& &  & %
    {\color[HTML]{999999} 64.75} & {\color[HTML]{999999} 67.84} & {\color[HTML]{999999} ---} & {\color[HTML]{999999} 49.73} & {\color[HTML]{999999} 62.66} & {\color[HTML]{999999} \emph{51.18}} & {\color[HTML]{999999} 61.01} & {\color[HTML]{999999} 83.38} & {\color[HTML]{999999} ---} \\%
\cmidrule{2-12}
& \textbf{EMSANet} & \textbf{2x ResNet101} & %
    63.63 & 66.36 & --- & 44.63 & 61.92 & \emph{50.35} & 59.82 & 83.51 & --- \\
& &  & %
    {\color[HTML]{9B9B9B} 64.11} & {\color[HTML]{9B9B9B} 66.64} & {\color[HTML]{9B9B9B} ---} & {\color[HTML]{9B9B9B} 46.32} & {\color[HTML]{9B9B9B} 61.96} & {\color[HTML]{9B9B9B} \emph{50.35}} & {\color[HTML]{9B9B9B} 59.82} & {\color[HTML]{9B9B9B} 83.80} & {\color[HTML]{9B9B9B} ---} \\%
\cmidrule{3-12}
& & \textbf{2x ResNet34-NBt1D} & %
    61.25 & 65.57 & --- & 45.47 & 58.32 & \emph{47.76} & 56.85 & 83.39 & --- \\
& &  & %
    {\color[HTML]{999999} 61.25} & {\color[HTML]{999999} 65.57} & {\color[HTML]{999999} ---} & {\color[HTML]{999999} 46.35} & {\color[HTML]{999999} 58.32} & {\color[HTML]{999999} \emph{47.76}} & {\color[HTML]{999999} 56.85} & {\color[HTML]{999999} 83.47} & {\color[HTML]{999999} ---} \\%
\bottomrule
\end{tabular}
}
\label{tab:results_sunrgbd_scannet2}
\vspace{-4mm}
\end{table*}
Even with a lower power profile~(measured power consumption of 30$\,$W), our proposed multi-task EMSAFormer approach meets our real-time requirements of at least 20$\,$FPS.

\begin{figure}[!b]%
    \centering%
    \vspace{-5mm}%
    \resizebox{\linewidth}{!}{%
    \begin{tikzpicture}%
        \tikzstyle{scene_label}=[anchor=south east, text=white, fill=black, fill opacity=0.3, text opacity=1, inner sep=2pt, rounded corners=2pt, minimum height=3.5mm, font=\tiny]
        \node[anchor=north] at (0, -4.6){\footnotesize{NYUv2 (test)}};%
        \node[anchor=north] at (0, 0){%
	        \includegraphics[width=2.9cm, height=2.175cm]{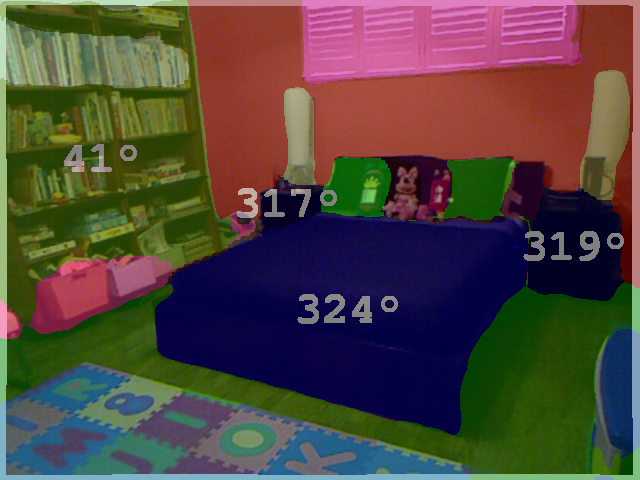}%
	    };%
	    \node[scene_label] at (1.45, -2.3){bedroom};%
	    \node[anchor=north] at (3, -4.6){\footnotesize{SUNRGB-D (test)}};%
	    \node[anchor=north] at (3, 0){%
	        \includegraphics[width=2.9cm, height=2.175cm]{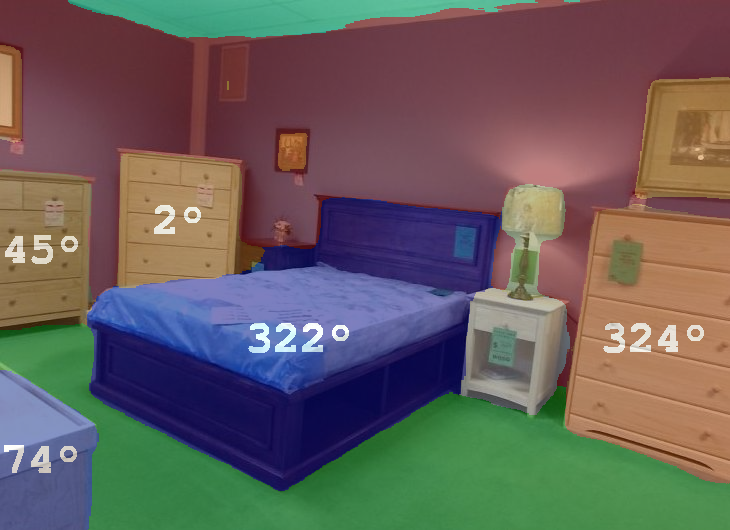}%
	    };%
	    \node[scene_label] at (4.45, -2.3){bedroom};%
	    \node[anchor=north] at (6, -4.6){\footnotesize{ScanNet (validation)}};%
	    \node[anchor=north] at (6, 0){%
	        \includegraphics[width=2.9cm, height=2.175cm]{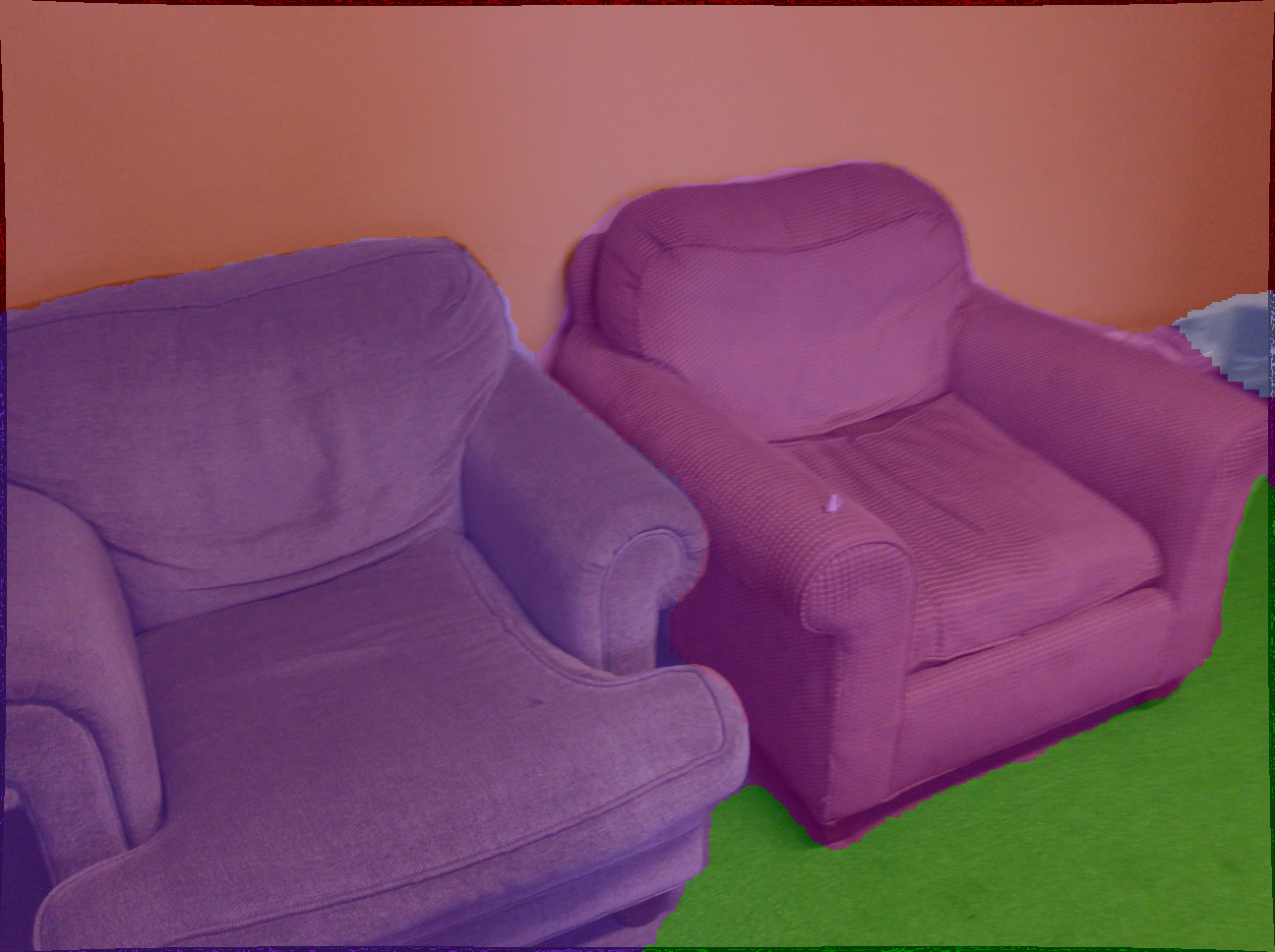}%
	    };%
	    \node[scene_label] at (7.45, -2.3){living room};%
	    \node[anchor=north] at (0, -2.3){%
	        \includegraphics[width=2.9cm, height=2.175cm]{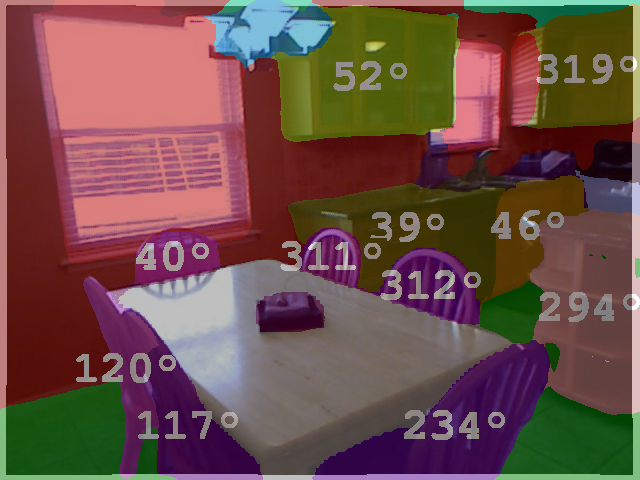}%
	    };%
	    \node[scene_label] at (1.45, -4.6){dining room};%
	    \node[anchor=north] at (3, -2.3){%
	        \includegraphics[width=2.9cm, height=2.175cm]{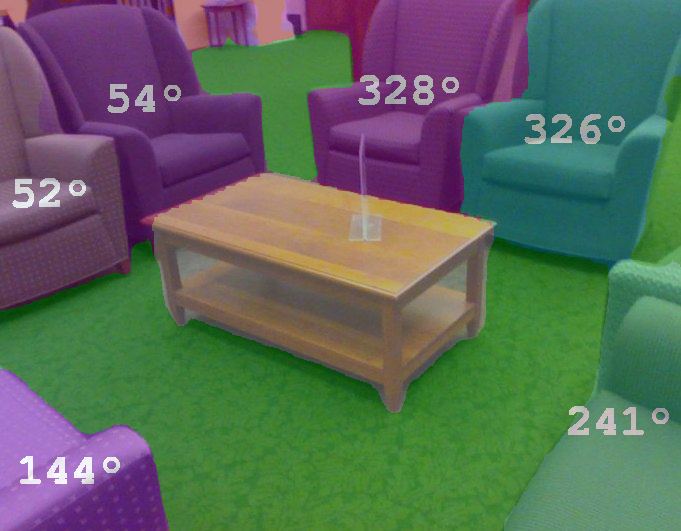}%
	    };%
	    \node[scene_label] at (4.45, -4.6){discussion room};%
	    \node[anchor=north] at (6, -2.3){%
	        \includegraphics[width=2.9cm, height=2.175cm]{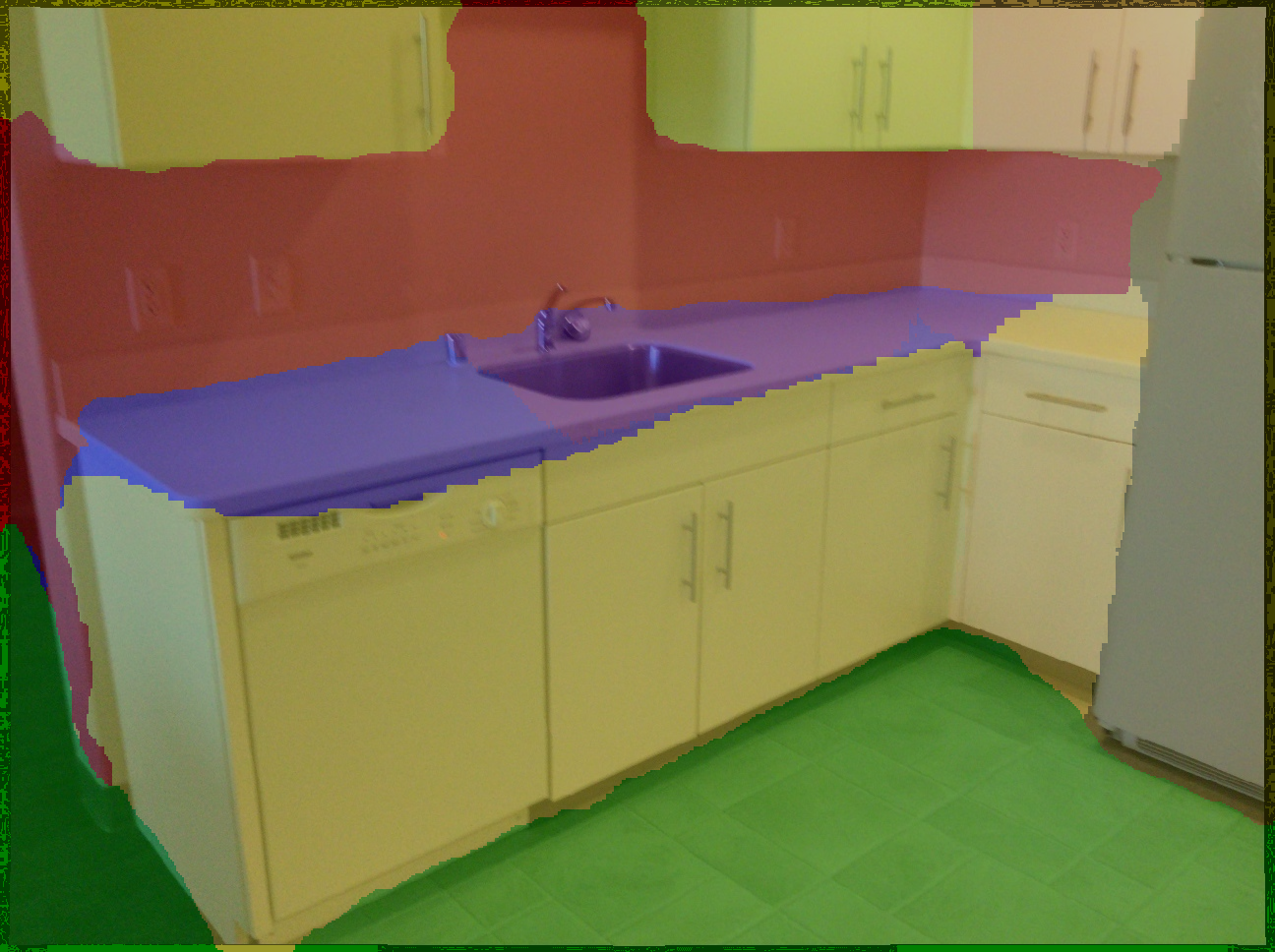}%
	    };%
	    \node[scene_label] at (7.45, -4.6){kitchen};%
    \end{tikzpicture}%
    }%
    \vspace{-4mm}%
    \caption{%
        Qualitative results as RGB image overlayed with predicted panoptic segmentation, predicted scene class, and estimated orientations if available.
    }
    \vspace{-0.5mm}%
    \label{fig:qualitative_results}
\end{figure}
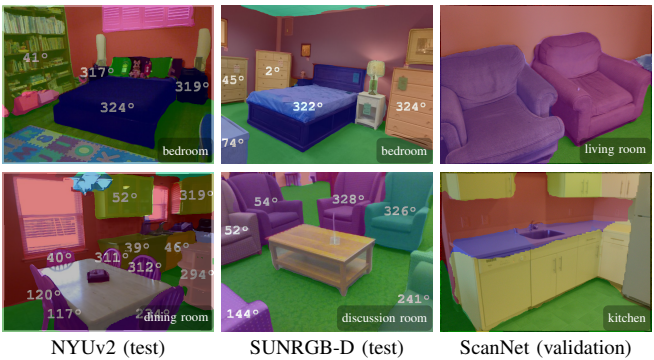

Fig.~\ref{fig:qualitative_results} presents qualitative results.
For all indoor datasets, our approach is able to analyze the scenes thoroughly.
The obtained predictions are well suited for enabling a mobile robot to operate autonomously in indoor environments.

\subsection{Comparison to State of the Art}
\label{sec:experiments:comparison_to_sota}
Comparing our proposed EMSAFormer to other approaches is challenging, as they mainly focus on single-task semantic segmentation and rarely account for efficiency.
Moreover, most approaches use test-time augmentation, which is not applicable on a mobile robot with limited computational resources.
The only approach that fits our multi-task setting is EMSANet~\cite{emsanet2022ijcnn}, which we already compared to above.
However, Tab.~\ref{tab:results_sota} shows additional comparisons to RGB-D approaches on all three common indoor datasets for semantic segmentation.
The results on NYUv2 and SUNRGB-D reveal that our approach outperforms other CNN-based approaches.
For NYUv2, the results are close to the OMNIVORE approach with the larger Swin-B backbone.
We also report results for the official ScanNet benchmark~(hidden test split).
Our approach outperforms EMSANet with a dual ResNet101 encoder on this split as well.
\section{Conclusion and Future Work}
\label{sec:conclusion}
We have presented a Transformer-based RGB-D approach for multi-task scene analysis, called EMSAFormer, that simultaneously performs panoptic segmentation, instance orientation estimation, and scene classification.
We have shown that the CNN-based dual encoder of EMSANet~\cite{emsanet2022ijcnn} can be replaced with a single Trasformer-based encoder.
Our extensive experiments on the three common indoor datasets NYUv2, SUNRGB-D, and ScanNet highlight the strong performance of our proposed EMSAFormer.
We further have revealed limitations of Transformer-based approaches in a single-task setting on smaller datasets, such as NYUv2.
However, we have demonstrated that these issues can be addressed using a multi-task approach.
Due to the proposed NVIDIA TensorRT extension, our EMSAFormer approaches can be applied in real time with 39.1$\,$FPS on an NVIDIA Jetson AGX Orin 32$\,$GB, demonstrating its suitability for deployment to mobile robots.
In future work, we intend to explore the benefits of additional training on large-scale RGB-D datasets such as Hypersim~\cite{hypersim-iccv2021}.

\begin{table}[!b]
\vspace{-5.5mm}
\centering
\caption{%
    Comparison to other state-of-the-art approaches on NYUv2 test split, SUNRGB-D test split, and ScanNet test~(benchmark) split. %
    * indicates additional test-time augmentation.
}
\vspace{-3mm}
\scriptsize%
\scalebox{0.91}{%
\begin{tabular}{@{\hspace{1mm}}l@{\hspace{2.5mm}}lll@{\hspace{1mm}}}
\toprule
 &  & \textbf{Backbone} & \multicolumn{1}{@{}c@{}}{\textbf{mIoU}$\,{}^\uparrow$} \\
\midrule
\multirow{8}{*}{\rotatebox{90}{\color[HTML]{9B9B9B}\textbf{NYUv2}}} 
                               & OMNIVORE~\cite{omnivore-cvpr2022}                & Swin-T                        & 47.9          \\
                               &                                                  & Swin-B                        & 51.1         \\
\cmidrule{2-4}
                               & ShapeConv~\cite{shapeconv-iccv2021}              & ResNet50                      & 47.3          \\
                               &                                                  & ResNext101                    & 50.2          \\
\cmidrule{2-4}                               
                               & SA-Gate~\cite{SA-Gate-eccv2020}                  & 2xResNet50                    & 50.4          \\
\cmidrule{2-4}
                               & \textbf{EMSAFormer (ours)}                       & SwinV2-T-128-Multi-Aug        & 51.26         \\
\midrule%
\multirow{5}{*}{\rotatebox{90}{\color[HTML]{9B9B9B}\textbf{SUNRGB-D}~$\,$}}
                               & ShapeConv~\cite{shapeconv-iccv2021}              & ResNet101                     & 47.6          \\
                               & AC-Net~\cite{ACNet-icip2019}                     & 3x ResNet50                   & 48.1          \\
                               & 2.5D Conv~\cite{2.5DConvolutionRGBD-icip2019}    & ResNet-101                    & 48.2          \\
                               & ESANet~\cite{esanet2021icra}                     & 2x ResNet50                   & 48.31         \\
\cmidrule{2-4}                              
                               & \textbf{EMSAFormer (ours)}                       & SwinV2-T-128-Multi-Aug (Sem(SegFormer)) & 48.82         \\
\midrule%
\multirow{4}{*}{\rotatebox{90}{\color[HTML]{9B9B9B}\textbf{ScanNet}~$\;$}}
                               & FuseNet~(from \cite{SSMA-ijcv2019})              & 2x VGG16                      & 53.5          \\
                               & SSMA~\cite{SSMA-ijcv2019}                        & 2x mod. ResNet50              & 57.7*         \\
                               & EMSANet                                          & 2x ResNet101                  & 54.0          \\
\cmidrule{2-4}
                               & \textbf{EMSAFormer (ours)}                       & SwinV2-T-128-Multi-Aug (Sem(SegFormer))  & 56.4          \\
\bottomrule
\end{tabular}
}
\label{tab:results_sota}
\end{table}
\bibliographystyle{IEEEtran}
\bibliography{bib/literature.bib}

\end{document}